\documentclass[conference]{IEEEtran}

\usepackage{times}
\usepackage{soul}
\usepackage{url}
\usepackage[hidelinks]{hyperref}
\usepackage[utf8]{inputenc}
\usepackage[small]{caption}
\usepackage{graphicx}
\usepackage{amsmath}
\usepackage{amsthm}
\usepackage{booktabs}
\usepackage{algorithmic}
\usepackage{multicol}
\usepackage{amsmath, multicol, multirow, booktabs, microtype, balance}
\usepackage[switch]{lineno}
\usepackage{xcolor}
\usepackage{subfig} 
\definecolor{amber}{rgb}{1.0, 0.6, 0.0}
\def\BibTeX{{\rm B\kern-.05em{\sc i\kern-.025em b}\kern-.08em
    T\kern-.1667em\lower.7ex\hbox{E}\kern-.125emX}}
\begin{document}

\title{MambaTab: A Plug-and-Play Model for Learning Tabular Data}

\author{
    \IEEEauthorblockN{Md Atik Ahamed\IEEEauthorrefmark{1}, Qiang Cheng\IEEEauthorrefmark{1}\IEEEauthorrefmark{2}\IEEEauthorrefmark{3}}
    \IEEEauthorblockA{\IEEEauthorrefmark{1}Department of Computer Science, 
    \IEEEauthorrefmark{2}Institute for Biomedical Informatics}
    \IEEEauthorblockA{University of Kentucky, Lexington, KY, USA}
    \{atikahamed, qiang.cheng\}@uky.edu \\
    \IEEEauthorrefmark{3}Corresponding author}

\maketitle

\begin{abstract}
Despite the prevalence of images and texts in machine learning, tabular data remains widely used across various domains. Existing deep learning models, such as convolutional neural networks and transformers, perform well however demand extensive preprocessing and tuning limiting accessibility and scalability. This work introduces an innovative approach based on a structured state-space model (SSM), MambaTab, for tabular data. SSMs have strong capabilities for efficiently extracting effective representations from data with long-range dependencies. MambaTab leverages Mamba, an emerging SSM variant, for end-to-end supervised learning on tables. Compared to state-of-the-art baselines, MambaTab delivers superior performance while requiring significantly fewer parameters, as empirically validated on diverse benchmark datasets. MambaTab’s efficiency, scalability, generalizability, and predictive gains signify it as a lightweight, “plug-and-play” solution for diverse tabular data with promise for enabling wider practical applications.
\end{abstract}

\section{Introduction}
Tabular data, with its structured format, is widely used in industrial, healthcare, academic, and other domains, despite the rise of image and Natural Language Processing (NLP) techniques in machine learning (ML). Numerous ML strategies, including traditional models and newer deep learning (DL) architectures like multi-layer perceptron (MLP), convolutional neural networks (CNNs), and Transformers \cite{vaswani2017attention}, have been adapted for tabular data, providing valuable insights and analytics.

State-of-the-art deep tabular models either have limited performance or require many parameters, extensive preprocessing, and tuning, using substantial resources, thus limiting their deployment \cite{wang2022transtab}. 
Furthermore, most methods require consistent table structures for training and testing and struggle with feature incremental learning, which involves sequentially adding features. This typically necessitates dropping either new features or old data, leading to an underutilization of available information \cite{wang2022transtab}. A model capable of continuous learning from new features is needed.

To address these challenges, we propose a new approach for tabular data based on structured state-space models (SSMs) \cite{gu2021combining,gu2021efficiently,fu2022hungry}. These models can be interpreted as a combination of CNNs and recursive neural networks, having advantages of both types of models. They offer parameter efficiency, scalability, and strong capabilities for learning representations from varied data, particularly for sequential data with long-range dependencies. To tap into these potential advantages, we leverage SSMs as an alternative to CNNs or Transformers for modeling tabular data. 

Specifically, we leverage Mamba \cite{gu2023mamba}, an emerging SSM variant, as a critical building block to build a novel model called \emph{MambaTab}. 

This proposed model has several key advantages over existing models: It not only requires significantly fewer model weights and exhibits linear parameter growth but also inherently aligns well with feature incremental learning. Additionally, MambaTab has a simple architecture that needs minimal data preprocessing. Finally, MambaTab outperforms state-of-the-art baselines, including MLP-, Transformer-, and CNN- based models and classic ML models. 

We benchmark MambaTab extensively against leading tabular data models.  Experiments under three different settings - vanilla supervised learning, self-supervised learning, and feature incremental learning - on 8 public datasets demonstrate MambaTab's superior performance. It consistently and significantly outperforms the state-of-the-art baselines, including Transformer-based models, while using a small fraction, typically \emph{$\mathbf{<1\%}$}, of their parameters.

In summary, the key innovations and contributions of MambaTab are:
\begin{itemize}
    \item Extremely small model size and number of learning parameters
    \item Linear scalability of model parameters in Mamba blocks, number of features, or sequence length
    \item Effective end-to-end training and inference with minimal data wrangling needed.
    \item Superior performance over state-of-the-art tabular learning approaches
\end{itemize}
As the first Mamba-based architecture for tabular data, MambaTab's advantages suggest that it can serve as an \emph{out-of-the-box}, plug-and-play model for tabular data on systems with varying computational resources. This holds promise to enable wide applicability across diverse practical settings.

\section{Related Work}
\label{sec:literature}
We review existing tabular data learning approaches, roughly categorizing them into classical models, deep learning (CNN or Transformer-based), and self-supervised strategies.

\noindent{{\bf{Classic Learning-based Approaches:}}} A variety of models exist based on classic ML techniques such as logistic regression (LR), XGBoost \cite{chen2016xgboost} \cite{zhang2020customer}, and MLP. Notably, a self-normalizing neural network (SNN) \cite{klambauer2017self}, an MLP variant using the scaled exponential linear unit (SELU), is tailored for tabular data. SNNs stabilize neuron activations at zero mean and unit variance, facilitating high-level abstract representations.

\noindent{{\bf{Deep Learning-based Supervised Models:}}} TabNet \cite{arik2021tabnet} is a DL model that employs an attention mechanism for tabular data, focusing on the most salient features at each decision step for efficient learning and interpretability. It performs well on various tabular datasets and provides interpretable feature attributions. Deep cross networks (DCN) \cite{wang2017deep} combine a deep network for learning high-order feature interactions and a cross-network that automatically applies feature crossing via a vector-wise cross operation. DCN efficiently learns bounded-degree feature interactions. 

A variety of models have been developed with Transformers as building blocks. 
AutoInt \cite{song2019autoint} uses Transformers to learn the importance of different input features. By relying on self-attention networks, this model can automatically learn high-order feature interactions in a data-driven way. TabTransformer \cite{huang2020tabtransformer} is also built upon self-attention based Transformers, which transform the embeddings of categorical features into robust contextual embeddings to achieve higher prediction accuracy. The contextual embeddings are shown to be highly robust against both missing and noisy data features and provide better interpretability. Moreover, FT-Transformer \cite{fttrans} tokenizes categorical features into continuous embeddings and models their interactions using Transformers.

\noindent{{\bf{Self-Supervised Learning-based Models}}} Several approaches for pre-training deep learning models using self-supervised strategies have emerged. VIME \cite{yoon2020vime} uses tabular data augmentation for self- and semi-supervised learning, creating pretext tasks to estimate mask vectors from corrupted data and for data reconstruction. SCARF \cite{bahri2021scarf} employs a self-supervised contrastive learning technique for tabular datasets, generating views for learning by corrupting random feature subsets.
TransTab \cite{wang2022transtab} proposes a novel framework for learning from tabular data across tables with different learning strategies including  two pre-training
strategies of Vertical-Partition Contrastive Learning (VPCL)
via supervised and self-supervised techniques and feature incremental learning. UniTabE \cite{yang2024unitabe}
pretrains a large network for tabular data using a masked language modeling strategy. Transformer is used as an integral component in both TransTab and UniTabE models.

\section{Method}
\subsection{Preliminaries}
Inside a Mamba block, two fully connected layers in two branches calculate linear projections ($LP_1, LP_2)$. The first branch $LP_1$'s output passes through a 1D causal convolution and SiLU activation  ${\mathcal{S}}(\cdot)$~\cite{silu}, then a structured state space model (SSM). The continuous-time SSM is a system of first-order ordinary differential equation mapping an input function or sequence $u(t)$ to output $x(t)$ through a latent state $h(t)$: 
\begin{equation}
    \label{eq:CT-SSM}
    dh(t)/dt = A \ h(t) + B \  u(t), \quad x(t) = C \ h(t), 
\end{equation}
where $h(t)$ is $N$-dimensional, with $N$ also known as a state expansion factor, $u(t)$ is $D$-dimensional, with $D$ being the $Dimension$ factor or the number of channels, $x(t)$ is usually taken as $D$ dimensional, and $A$, $B$, and $C$ are coefficient matrices of appropriate sizes. This dynamic system induces a discrete version governing state evolution and SSM outputs given the input token sequence via time sampling at $\{ k \Delta \}$ with a $\Delta$ time interval. This discrete SSM version is a difference equation:
\begin{equation}
    \label{eq:DT-SSM}
    h_k = \bar{A} \ h_{k-1} + \bar{B} \ u_{k}, \quad x_k = C \ h_{k}, 
\end{equation}
where $h_k$, $u_k$, and $x_k$ are respectively samples of $h(t)$, $u(t)$, and $x(t)$ at time $k \Delta$, $\bar{A} = \exp(\Delta A)$, and $\bar{B} = (\Delta A)^{-1} (\exp(\Delta A) - I) \Delta B$. For SSMs, diagonal $A$ is often used, and Mamba also makes $B$, $C$, and $\Delta$ linear time-varying functions dependent on the input. With such time-varying coefficient matrices, the resulting SSM possesses context and input selectivity properties \cite{gu2023mamba}, facilitating Mamba blocks to selectively propagate or forget information along the potentially long input token sequence based on the current token. Subsequently, the SSM output is multiplicatively modulated with $\mathcal{S}(LP_2)$ before another fully connected projection.
\subsection{Architecture of Our Model}
\begin{figure*}[t]
    \centering
    \includegraphics[width=0.9\linewidth]{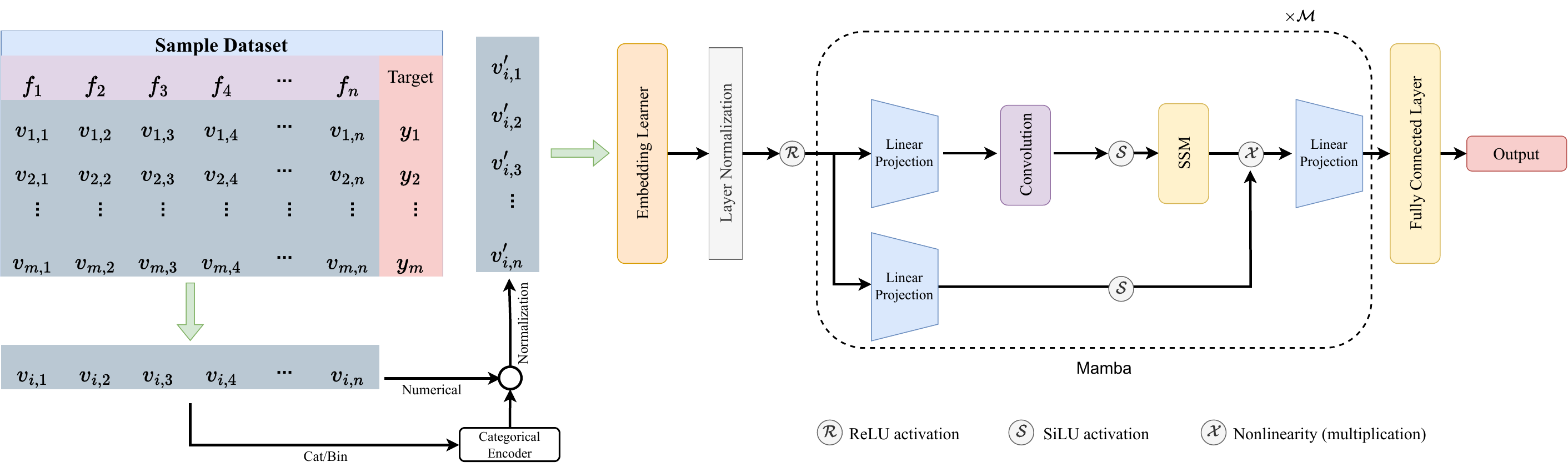}
    \caption{Schematic diagram of our proposed method (MambaTab). \textbf{Left:} Data preprocessing and representation learning. The embedding learner module is critical to ensure the embedded feature dimension is the same before and after new features are added under incremental learning. \textbf{Right:} Conversion of input data to prediction values via Mamba and a fully connected layer.}
    \label{fig:method}
\end{figure*}

\begin{figure}
    \centering
    \includegraphics[width=0.7\linewidth]{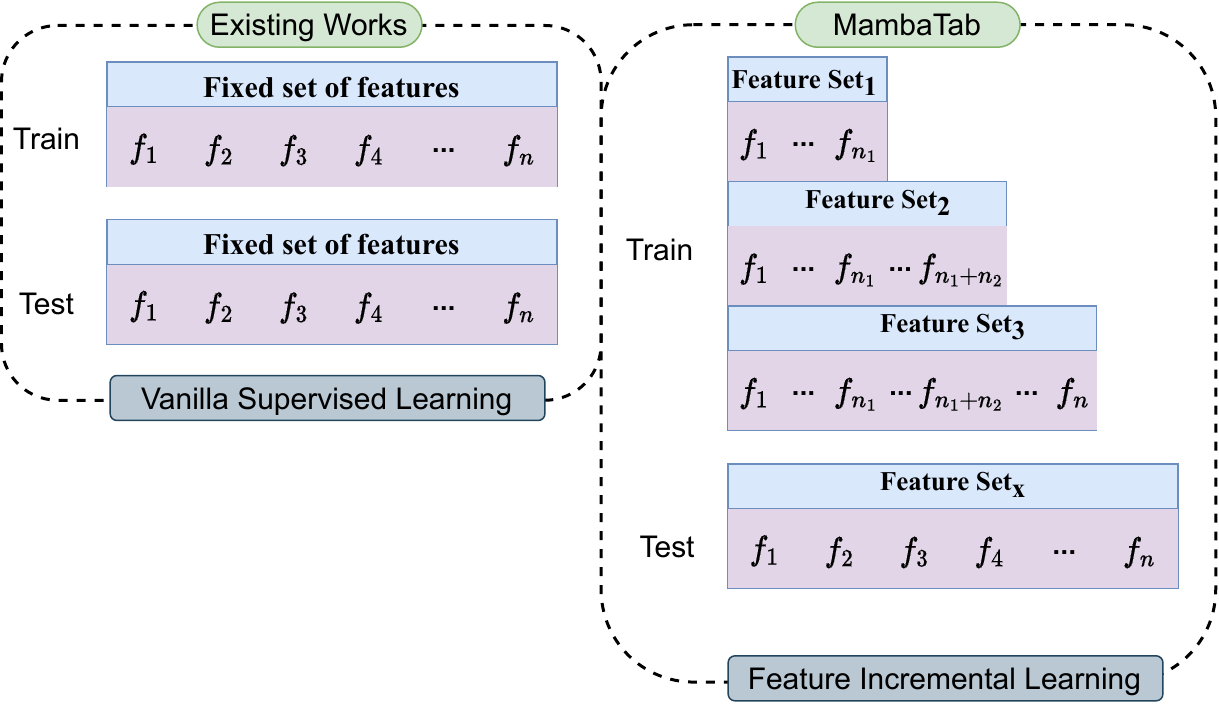}
    \caption{Illustration of feature incremental learning setting. Feature Set${_i}$, $i = 1,2,3$, have incrementally added features. Feature Set${_X}$ represents the set of features for test data.}
    \label{fig:method_incremental}
\end{figure}

In this section, we present our approach for robust learning of tabular data classification, aiming to improve performance through a plug-and-play, efficient, yet effective method. Below we describe each component of our method.\\
\noindent{\bf{Data preprocessing:}}
\label{sec:method_preprocess}
We consider a tabular dataset, $\{F_i,y_i\}_{i=1}^m$, where the features of the $i$-th sample are represented by $F_i=\{v_{i,j}\}_{j=1}^n$, its corresponding label is $y_i\in \{ 0,1 \}$, and $v_{i,j}$ can be categorical, binary or numerical. We treat both binary and categorical features as categorical and utilize an ordinal encoder for encoding them, as shown in Figure~\ref{fig:method}. Unlike TransTab~\cite{wang2022transtab}, our method does not require manual identification of feature types such as categorical, numerical, or binary. Moreover, MambaTab only requires 1 embedding learner module whereas TransTab requires 4 embedding modules.
We keep numerical features unchanged in the dataset and handle missing values by imputing the mode.  This preprocessing preserves the feature set cardinality, i.e. $n(F_i)=n({F_i}')$, where $n(F_i)$ 
and $n({F_i}')$ are the numbers of features before and after processing. Before feeding data into our model, we normalize 
values $v_{i,j}\in [0,1]$ using min-max scaling.

\noindent{\bf{Embedded representation learning:}}
With processed data, we employ a feed-forward network to learn an embedded representation from the features, providing meaningful inputs to our architecture. Although the ordinal encoder imposes ordered representations for categorical features, not all inherently possess such order. Our embedding learner allows the direct learning of multi-dimensional representations from features without depending on imposed orders. This approach also standardizes input feature dimensions for the downstream Mamba blocks across both training and testing in incremental feature learning scenarios, as shown in Figure~\ref{fig:method_incremental}. While most existing methods, except TransTab, are only capable of learning from a fixed set of features, our method MambaTab can learn and transfer weights from Feature Set$_1$ to Feature Set$_2$, and so on. We also utilize layer normalization~\cite{ba2016layer} instead of batch normalization~\cite{ioffe2015batch} on the learned embedded representations due to its independence of batch size.\\
\noindent{\bf{Cascading Mamba Blocks:}}
After getting the normalized embedded representations from layer normalization, we apply ReLU activation~\cite{agarap2018relu} and pass the resulting values $\{u_k^i\}$, with $u_k^i$ being the $k$-th token for example $i$, to a Mamba block~\cite{gu2023mamba}. This map features $Batch \times Length \times Dimension \rightarrow Batch \times Length \times Dimension$. Here, $Batch$ is the minibatch size; $Length$ refers to the token sequence length, and $Dimension$ is the number of channels for each input token. For simplicity, we use $Length=1$ by default and $Dimension$ matches the output dimension from the embedding learning layer (Figure~\ref{fig:method}). Although Mamba blocks can repeat ${\mathcal{M}}$ times, we set ${\mathcal{M}}=1$ as our default value. However, we perform a sensitivity study for ${\mathcal{M}}=2, \cdots, 100$ with stacked Mamba blocks, which are connected with residual connections~\cite{he2016deep}, to evaluate their information retention or propagation capacity. As a result, these integrated blocks empower MambaTab for content-dependent feature extraction and reasoning with long-range dependencies and feature interactions.\\ 
\noindent{\bf{Output Prediction:}}
In this portion, our method learns representations from the concatenated Mamba blocks' output $\{x_k^i\}$ of shape $Batch \times Length \times Dimension$, where $x_k^i$ is the $k$-th token output for example $i$ in a minibatch. These are projected via a fully connected layer from $Batch \times Length \times Dimension \rightarrow Batch \times 1$, resulting in prediction logit $y'_i$ for example $i$. With sigmoid activation, we obtain the predicted probability score for calculating AUROC and binary-cross-entropy loss. 

\section{Experiments}
\label{sec:experiments}
\subsection{Datasets, Implementation Details, and Baselines}
{\noindent\bf{Datasets:}} To systematically evaluate the effectiveness of our method, we utilize 8 diverse public benchmark datasets which are also widely followed by TransTab~\cite{wang2022transtab} and UniTabE~\cite{yang2024unitabe}. We provide the dataset's details and abbreviations in Table~\ref{tab:dataset_details}. Our default experimental settings follow those of ~\cite{wang2022transtab}. We split all datasets into train (70\%), validation (10\%), and test (20\%).

\noindent{{\bf{Implementation Details:}}} To keep the preprocessing simple, we follow the approach described in Section~\ref{sec:method_preprocess}, generalizing for all datasets without manual intervention. For post-training validation, we take the best validation model and use it on the test set for prediction. We set up MambaTab with default hyperparameters and tuned potential hyperparameters for each dataset under vanilla supervised learning. For our default hyperparameters, we set up training for 1000 epochs with early stopping patience = 5. We adopt Adam optimizer~\cite{kingma2014adam} and cosine-annealing learning rate scheduler with initial learning rate = $1e^{-4}$. In addition to training hyperparameters, MambaTab also involves other model-related hyperparameters and their default values are: embedded representation size = 32, SSM state expansion factor ($N$) = 32, local convolution width (d\_conv) = 4, number of SSM blocks ($\mathcal{M}$) = 1. 

\noindent{{\bf{Baselines:}}} We extensively benchmark our model by comparing it against standard and current state-of-the-art methods. These include: LR, XGBoost, MLP, SNN with SELU MLP, TabNet, DCN, AutoInt, TabTransformer, FT-Transformer, VIME, SCARF, UniTabE and TransTab. More information about them can be found in Section \ref{sec:literature}. For a fair comparison, we follow their implementation detailed in TransTab \cite{wang2022transtab}. 

\noindent{{\bf{Performance Benchmark:}}} With default hyperparameters under vanilla supervised learning, our method denoted by MambaTab-D achieves better performance than state-of-the-art baselines on many datasets and comparable performance on others with far fewer parameters (Table~\ref{tab:pub_datasets_params_sizes}). After tuning hyperparameters, we denote our tuned model by MambaTab-T, whose performance further improves. Moreover, under feature incremental learning, our method substantially outperforms the existing method simply with default hyperparameters. We implement MambaTab in PyTorch, which can be found here\footnote{https://github.com/Atik-Ahamed/MambaTab}. For evaluation, we use  Area Under the Receiver Operating Characteristic (AUROC) following ~\cite{fttrans,wang2022transtab,yang2024unitabe}.

\begin{table}[t]
\centering
\caption{Publicly available datasets with statistics.}
\label{tab:dataset_details}
\resizebox{\linewidth}{!}{
\begin{tabular}{lcccccc}

\toprule
    Dataset Name & Abbreviation & Datapoints & Train & Val & Test & Positive\\ 
\midrule
\href{https://openml.org/search?type=data\&status=active\&id=31}{Credit-g}  & CG & 1000 & 700& 100& 200& 0.70\\
\href{https://archive.ics.uci.edu/ml/datasets/credit+approval}{Credit-approval} & CA & 690 & 483 & 69 & 138 & 0.56\\
\href{https://www.openml.org/search?type=data\&status=active\&id=23381}{Dresses-sales} & DS & 500 & 350 & 50 & 100 & 0.42\\
\href{https://www.openml.org/search?type=data\&status=active\&id=1590}{Adult} & AD & 48842 & 34189 & 4884 & 9769 & 0.24\\
\href{https://www.openml.org/search?type=data\&status=active\&id=6332}{Cylinder-bands}  & CB & 540 & 378 & 54 & 108 & 0.58\\
\href{https://www.kaggle.com/datasets/blastchar/telco-customer-churn}{Blastchar} & BL & 7043 & 4930 & 704 & 1409 & 0.27\\
\href{https://archive.ics.uci.edu/ml/datasets/Insurance+Company+Benchmark+\%28COIL+2000\%29}{Insurance-co} & IO & 5822 & 4075 & 582 & 1165 & 0.06\\
\href{https://www.kaggle.com/datasets/lodetomasi1995/income-classification}{Income-1995} & IC & 32561 & 22792 & 3256 & 6513 & 0.24\\
\bottomrule
\end{tabular}
}
\end{table}

\subsection{Vanilla Supervised Learning Performance}
For this setting, we follow the protocols from~\cite{wang2022transtab} directly using the training-validation sets for model learning-tuning and the test set for evaluation. To overcome potential sampling bias, we report average results over 10 runs with different random seeds on each of the 8 datasets. With defaults, MambaTab-D outperforms baselines on 3 public datasets (CG, CA, BL) and has comparable performance to transformer-based baselines on others. For example, MambaTab-D outperforms TransTab~\cite{wang2022transtab} on 5 out of 8 datasets (CG, CA, DS, CB, BL). After tuning hyperparameters, MambaTab-T achieves even better performance, outperforming all baselines on 6 datasets and achieving the second best on the other 2 datasets. In Table~\ref{tab:pub_dataset_results}, we did not find performance on the IC dataset from the UniTabE paper. UniTabE-S and UniTabE-FT (Table~\ref{tab:ssl}) refer to the cases of training from scratch and utilizing the pre-trained weights and fine-tuning on the target datasets, respectively.
\begin{table}[t]
\centering
\caption{Test AUROC for vanilla supervised learning. The best results are shown in \textbf{bold} and the second best are shown in \underline{underlined}.}
\label{tab:pub_dataset_results}
\setlength{\tabcolsep}{2pt}
\medskip
\resizebox{\linewidth}{!}{
\begin{tabular}{lccccccccc} 
\toprule
\multirow{2}{*}{Methods} & \phantom{a} &  \multicolumn{8}{c}{Datasets}\\ 
\cmidrule{3-10} 
&\phantom{a}& CG & CA & DS & AD & CB & BL & IO & IC\\
\cmidrule{3-10}
LR & \phantom{a} & 0.720 & 0.836 & 0.557 & 0.851  & 0.748 & 0.801 & 0.769 & 0.860\\
XGBoost & \phantom{a} & 0.726 & 0.895 & 0.587 & 0.912 & \underline{0.892} & 0.821& 0.758 & \bf0.925\\
MLP & \phantom{a} & 0.643 & 0.832 & 0.568 & 0.904 & 0.613 & 0.832 & 0.779 & 0.893\\
SNN & \phantom{a} & 0.641 & 0.880 & 0.540 & 0.902 & 0.621 & 0.834 & 0.794 & 0.892\\
TabNet & \phantom{a} & 0.585 & 0.800 & 0.478 & 0.904 & 0.680 & 0.819 & 0.742 & 0.896\\
\midrule
DCN & \phantom{a}& 0.739 & 0.870 & \underline{0.674} & \underline{0.913} & 0.848 &  0.840 & 0.768 & 0.915\\
AutoInt & \phantom{a} & 0.744 & 0.866 & 0.672 & \underline{0.913} & 0.808 & 0.844 & 0.762 & 0.916\\
\midrule
TabTrans & \phantom{a} & 0.718 & 0.860 & 0.648 & \bf0.914 & 0.855 & 0.820 & 0.794 & 0.882\\
FT-Trans & \phantom{a} & 0.739 & 0.859 & 0.657 & \underline{0.913}& 0.862 & 0.841 & 0.793 & 0.915\\
\midrule
VIME & \phantom{a} & 0.735 & 0.852 & 0.485 & 0.912 & 0.769 & 0.837 & 0.786 & 0.908\\ 
SCARF & \phantom{a} & 0.733 & 0.861 & 0.663 & 0.911 & 0.719 & 0.833 & 0.758 & 0.905\\
TransTab & \phantom{a} & 0.768 & 0.881 & 0.643 & 0.907 & 0.851 & 0.845 & \bf0.822 & 0.919\\
UniTabE-S& \phantom{a}&0.760&0.930&0.620&0.910&0.850&0.840&0.740&---\\
\midrule

MambaTab-D & \phantom{a} &\underline{0.771} & \underline{0.954} & 0.643 & 0.906 & 0.862 & \underline{0.852} & 0.785 & 0.906\\
MambaTab-T & \phantom{a} &\bf0.801&\bf0.963&\bf0.681&\bf0.914&\bf0.896 & \bf0.854 & \underline{0.812} & \underline{0.920}\\
\bottomrule
\end{tabular}
}
\end{table}

\subsection{Feature Incremental Learning Performance}
For this setting, we divide the feature set $F$ of each dataset into three non-overlapping subsets $s_1,s_2,s_3$. $set_1$ contains $s_1$ features, $set_2$ contains $s_1,s_2$ features, and $set_3$ contains all features in $s_1,s_2,s_3$. While other baselines can only learn from either $set_1$ by dropping all incrementally added features (with respect to $s_1$) or $set_3$ by dropping old data, TransTab~\cite{wang2022transtab} and MambaTab can incrementally learn from $set_1$ to $set_2$ to $set_3$. In our method, we simply change the input feature cardinality $n(set_i)$ between settings, with the remaining architecture fixed. Our method works because Mamba has strong content and context selectivity for extrapolation and we keep the representation space dimension fixed, that is, independent of feature set cardinality $n(F)$. Thus, this demonstrates the adaptability and simplicity of our method for incremental environments. Even with default hyperparameters, MambaTab-D outperforms all baselines as shown in Table~\ref{tab:feature_incremental_performance}. We could not directly find the performance of UniTabE for the feature incremental learning setting. Here, we report the results averaged over 10 runs with different random seeds. Since it already achieves strong performance, we do not tune the hyperparameters further, although doing so could potentially improve performance.
\begin{table}[t]
\centering
\caption{Test AUROC for feature incremental learning. The best results are shown in \textbf{bold}.}
\label{tab:feature_incremental_performance}
\setlength{\tabcolsep}{2pt}
\medskip
\resizebox{\linewidth}{!}{
\begin{tabular}{lccccccccc} 
\toprule
\multirow{2}{*}{Methods} & \phantom{a} &  \multicolumn{8}{c}{Datasets} \\ 
\cmidrule{3-10}
&\phantom{a}& CG & CA & DS & AD & CB & BL & IO & IC\\
\cmidrule{3-10}
LR & \phantom{a}&  0.670 & 0.773 &  0.475 & 0.832 & 0.727 &  0.806 &  0.655 & 0.825 \\
XGBoost & \phantom{a} & 0.608 & 0.817 & 0.527 & 0.891 & 0.778  & 0.816 & 0.692 & 0.898\\
MLP & \phantom{a} &  0.586 & 0.676 & 0.516 & 0.890 & 0.631 & 0.825 & 0.626 & 0.885\\
SNN & \phantom{a} & 0.583 & 0.738 & 0.442 & 0.888 & 0.644 & 0.818 & 0.643 & 0.881\\
TabNet & \phantom{a} &  0.573 & 0.689 & 0.419 & 0.886 & 0.571 & 0.837 & 0.680 & 0.882\\
\midrule
DCN & \phantom{a} & 0.674 & 0.835 & 0.578 & 0.893 & 0.778 & 0.840 & 0.660 & 0.891\\
AutoInt & \phantom{a} &  0.671 & 0.825 & 0.563 & 0.893 & 0.769 & 0.836 & 0.676 & 0.887\\
\midrule
TabTrans & \phantom{a} & 0.653 & 0.732 & 0.584 & 0.856 & 0.784 & 0.792 & 0.674 & 0.828\\
FT-Trans & \phantom{a} & 0.662 & 0.824 & 0.626 & 0.892 & 0.768 & 0.840 & 0.645 & 0.889\\
\midrule
VIME & \phantom{a} & 0.621 & 0.697 & 0.571 & 0.892 & 0.769 & 0.803 & 0.683 & 0.881\\ 
SCARF & \phantom{a} & 0.651 & 0.753 & 0.556 & 0.891 & 0.703 & 0.829 & 0.680 & 0.887\\
TransTab & \phantom{a} & 0.741 & 0.879 & 0.665 & 0.894 & 0.791 & 0.841 & 0.739 & 0.897\\
\midrule
MambaTab-D & \phantom{a}&\bf0.787&\bf0.961&\bf0.669&\bf0.904&\bf0.860&\bf0.853& \bf0.783 &\bf0.908\\
\bottomrule
\end{tabular}
}
\end{table}
\subsection{Self-Supervised Leraning}
For self-supervised learning (SSL), we randomly corrupted 50\% features in each iteration with zeros and tried to reconstruct the original sequence with our model. To achieve this, we utilized $L_2$ loss for reconstruction and changed the output projection dimension to be matched with the number of features passed as input to our model. With this strategy, our model learns more impactful underlying meanings of the data. Here again, we only utilized the train-val set for self-supervised learning keeping the test set completely unseen to the model. Moreover, our SSL technique is completely unaware of the true class label of the feature and only utilizes raw features, demonstrating its robustness towards leveraging large-scale unlabeled data. Our performance compared against TransTab's two pre-training strategies of  Vertical-Partition Contrastive Learning (VPCL) via supervised and self-supervised techniques are shown in Table~\ref{tab:ssl}, which demonstrates our method's superiority over TransTab while performing SSL fashioned learning. Moreover, our method also demonstrates superior performance over UniTabE-FT in many cases.
\begin{table}[t]
\centering
\caption{Test AUROC results under different pre-training schemas.}
\label{tab:ssl}
\setlength{\tabcolsep}{2pt}
\medskip
\resizebox{\linewidth}{!}{
\begin{tabular}{lccccccccc} 
\toprule
\multirow{2}{*}{Methods} &\multirow{2}{*}{Schema} &  \multicolumn{8}{c}{Datasets}\\ 
\cmidrule{3-10} 
&\phantom{a}& CG & CA & DS & AD & CB & BL & IO & IC\\
\multirow{2}{*}{TransTab}&Self-VPCL&0.777&0.837&0.626&0.907&0.819&0.843&\bf0.823&\bf0.919\\
&VPCL&0.776&0.858&0.637&0.907&0.862&0.844&\underline{0.819}&\bf0.919\\
\midrule
UniTabE&FT&0.790&0.940&\bf0.660&\bf0.910&0.880&0.840&0.760&---\\
MambaTab&SSL&\bf0.804&\bf0.967&\underline{0.649}&\underline{0.909}&\bf0.880&\bf0.857&0.786&\underline{0.909}\\

\bottomrule
\end{tabular}
}
\end{table}
\subsection{Learnable Parameter Comparison}
Our method not only achieves superior performance compared to existing state-of-the-art methods, it is also memory and space efficient. We demonstrate our method's superiority in terms of learnable parameter size while comparing against transformer-based approaches in Table~\ref{tab:pub_datasets_params_sizes}. It is seen that our method (both MambaTab-D/T) achieves comparable or better performance than TransTab typically with  $<1\%$ of its learnable parameters. To evaluate learnable parameter size, we use the default settings specified in FT-Trans, TransTab, and TabTrans~\footnote{https://github.com/lucidrains/tab-transformer-pytorch}.
We also notice that, despite varying features, TransTab's model size remains unchanged. The most important tunable hyperparameters for MambaTab include the controllable block expansion factor, the state expansion factor ($N$), and the embedded representation space dimension. We perform sensitivity analysis on them in Section \ref{sec:sensitivity-ana} and also fine-tune them for each dataset. In addition, we conduct an ablation study for the normalization layer of our model. 
\begin{table}[t]
\centering
\caption{Total learnable parameters (M = million, K = thousand).}
\label{tab:pub_datasets_params_sizes}
\setlength{\tabcolsep}{2pt}
\medskip

\begin{tabular}{lccccccccc} 
\toprule
\multirow{2}{*}{Methods} & \phantom{a} &  \multicolumn{8}{c}{Datasets} \\ 
\cmidrule{3-10}
&\phantom{a}& CG & CA & DS & AD & CB & BL & IO & IC\\
\cmidrule{3-10}
TabTrans & \phantom{a} & 2.7M &1.2M&2.0M&1.2M&6.5M &3.4M&87.0M&1.0M\\
FT-Trans & \phantom{a} & 176K &176K &179K&178K&203K&176K&193K&177K\\
TransTab & \phantom{a} &4.2M&4.2M&4.2M&4.2M&4.2M&4.2M &4.2M&4.2M\\
\cmidrule{1-10}
MambaTab-D & \phantom{a} & \bf13K&\bf13K&13K&\bf13K&\bf14K& 13K&15K&13K \\
MambaTab-T & \phantom{a} & 50K & 38K &  \bf5K & 255K & 30K & \bf11K & \bf13K & \bf10K\\
\bottomrule
\end{tabular}

\end{table}

\subsection{Hyperparameter Tuning}
We tune critical hyperparameters using validation loss, ensuring the test set remains untapped until final testing with the best-performing settings. We have reported averaged test results over 10 runs with different random seeds with the tuned MambaTab (Table \ref{tab:pub_dataset_results}). Learnable parameter sizes of MambaTab-T are reported in Table~\ref{tab:pub_datasets_params_sizes}. Interestingly, MambaTab-T sometimes consumes fewer parameters than even MambaTab-D, e.g., on DS, BL, IO, and IC. We demonstrate key components of our tuned model MambaTab-T in Table~\ref{tab:hyper_params}, where the tuned values for these components are shown, with other training-related hyperparameters, such as training epochs and learning rate, at default values; see Implementation Details in Section~\ref{sec:experiments}. 
\begin{table}[t]
\centering
\caption{Hyperparameters of our tuned model, MambaTab-T.}
\label{tab:hyper_params}
\setlength{\tabcolsep}{2pt}
\medskip
\small
\begin{tabular}{lccccccccc} 
\toprule

Hyperparameters & \phantom{a} &  \multicolumn{8}{c}{Datasets} \\ 
\cmidrule{3-10}
&\phantom{a}& CG & CA & DS & AD & CB & BL & IO & IC\\
\cmidrule{3-10}
Embedding Representation Space & \phantom{a} & 64 & 32 & 16 & 64 & 32 & 16 & 16 & 32\\
\cmidrule{3-10}
State Expansion Factor & \phantom{a} & 16 & 64 & 32 & 64 & 8 & 4 & 8 & 64\\
\cmidrule{3-10}
Block Expansion Factor& \phantom{a} & 3 & 4 & 2 & 10 & 7 & 10 & 9 & 1\\

\bottomrule
\end{tabular}

\end{table}
\section{Hyperparameter Sensitivity and Ablation Study}
\label{sec:sensitivity-ana}
In this section, we demonstrate extensive sensitivity analyses and ablation experiments on MambaTab's most important hyperparameters using two randomly selected datasets: Cylinder-Bands (CB) and Credit - g (CG). We measure performance by changing each factor, including block expansion factor, state expansion factor, and embedding representation space dimension, while keeping $\mathcal{M}=1$ and other hyperparameters at default values as in MambaTab-D. We report results averaged over 10 runs with different random splits to overcome the potential bias due to randomness.
\subsection{Block Expansion Factor}
We experiment with block expansion factor (kernel size) $\{1,2,...,10\}$, keeping the other hyperparameters at default values as in  MambaTab-D. As seen in Figure~\ref{fig:ablation_state_block}, MambaTab's performance changes only slightly with different block expansion factors, with no clear or monotonic trends. Thus we set the default to 2, inspired by~\cite{gu2023mamba}, though tuning this parameter further could improve performance on some datasets.

\subsection{State Expansion Factor}

\begin{figure}[t]
    \centering
    \subfloat[Block Expansion Factor]{ \includegraphics[width=0.4\linewidth]{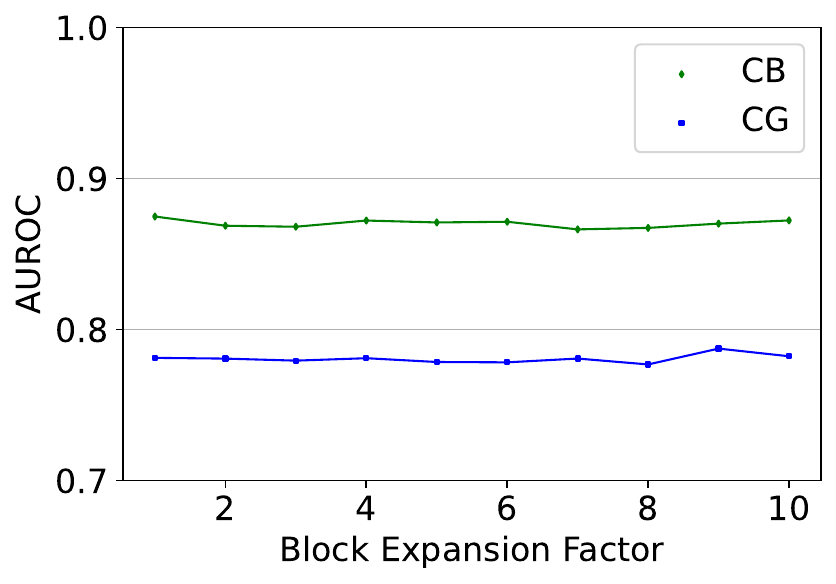}}%
    \qquad
    \subfloat[State Expansion Factor]{{\includegraphics[width=0.4\linewidth]{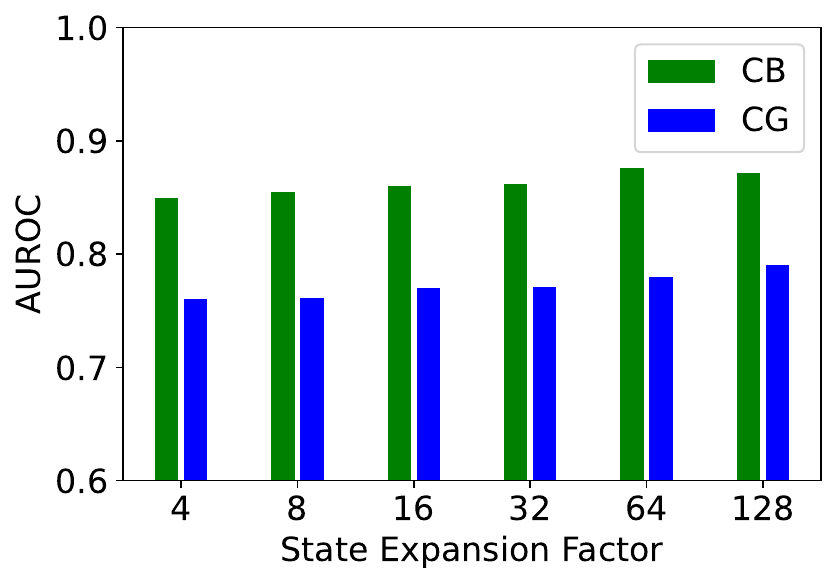} }}%
   
    \caption{Ablation on block and state expansion factors.}%
    \label{fig:ablation_state_block}
\end{figure}

We show the impact of the state expansion factor ($N$) using values in ${4,8,16,32,64,128}$, where MambaTab's AUROC improves with increasing $N$ for datasets CG and CB (Figure~\ref{fig:ablation_state_block}). Although a higher $N$ enhances performance, it also uses more memory. Balancing performance and memory use, we set 32 as the default $N$.

\subsection{Size of Embedded Representations}
\begin{figure}[t]
    \centering
    \includegraphics[width=0.4\linewidth]{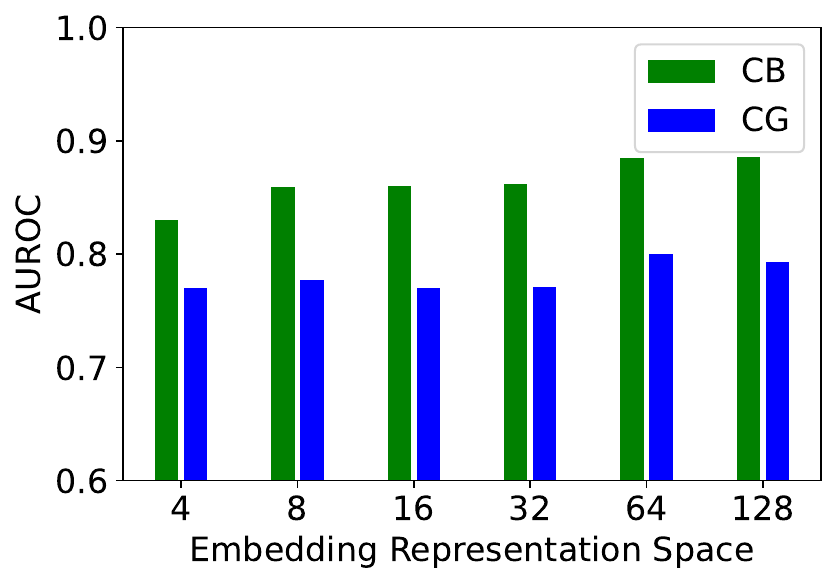}
    \caption{Ablation on embedded representation space.}
    \label{fig:ablation_representation}
\end{figure}

As mentioned in the Method section, we allow flexibility for the model to learn the embedding via a fully connected layer. We also perform sensitivity analysis for the length of the embedded representations, with values in $\{4, 8, 16, 32, 64, 128\}$. As seen in Figure~\ref{fig:ablation_representation}, MambaTab's performance
essentially increases for both CG and CB datasets with larger embedding sizes, though at the cost of more parameters and thus larger CPU/GPU space. To balance performance versus model size, we choose to keep the default embedding length to 32.

\subsection{Ablation of Layer Normalization}
We demonstrate the effect of layer normalization, which is applied to the embedded representations, in our model architecture shown in Figure~\ref{fig:method}. We contrast the performance by keeping or dropping this layer in vanilla supervised learning experiments on CG and CB datasets. The results in AUROC metric are shown in Table~\ref{tab:layer_norm}.
\begin{table}[t]
\centering
\caption{Ablation analysis of layer normalization via AUROC}
\label{tab:layer_norm}
\setlength{\tabcolsep}{2pt}
\medskip

\begin{tabular}{lcccc} 
\toprule
Ablation & \phantom{a} &  \multicolumn{3}{c}{Datasets} \\ 
\midrule
&& CG &\phantom{a}& CB\\
Without Layer Normalization & \phantom{a} & 0.759 & \phantom{a} & 0.847\\
With Layer Normalization &\phantom{a} & \bf0.771 & \phantom{a} & \bf0.862 \\

\bottomrule
\end{tabular}

\end{table}
Without layer normalization, the embeddings would directly pass through the ReLU activation, as shown in the overall scheme (Figure~\ref{fig:method}). On both CG and CB datasets, MambaTab's performance improves with layer normalization versus without.

\subsection{Scaling Mamba}
\begin{figure}[t]
    \centering
    \subfloat[AUROC results (CB)]{{\includegraphics[trim={0.25cm 0.25cm 0.3cm 0.25cm},clip,width=0.38\linewidth]{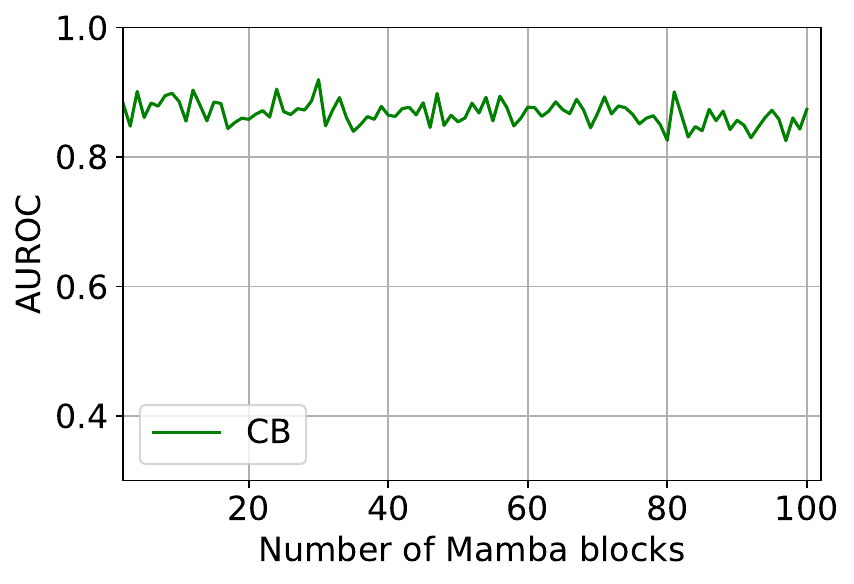} }}%
    \qquad
    \subfloat[Learnable parameters (CB)]{{\includegraphics[trim={0.25cm 0.25cm 0.3cm 0.25cm},clip,width=0.38\linewidth]{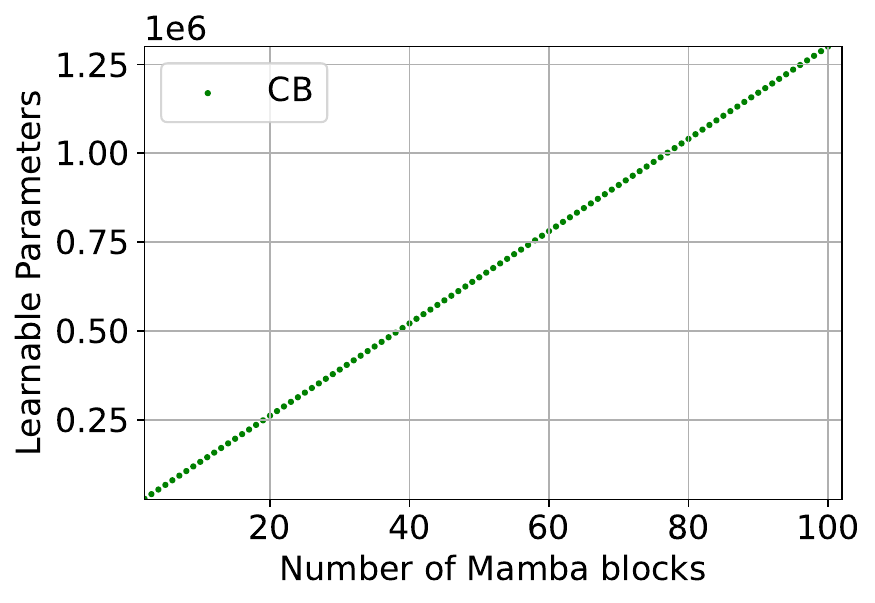} }}%
    \qquad
    \subfloat[AUROC results (CG)]{{\includegraphics[trim={0.25cm 0.25cm 0.3cm 0.25cm},clip,width=0.38\linewidth]{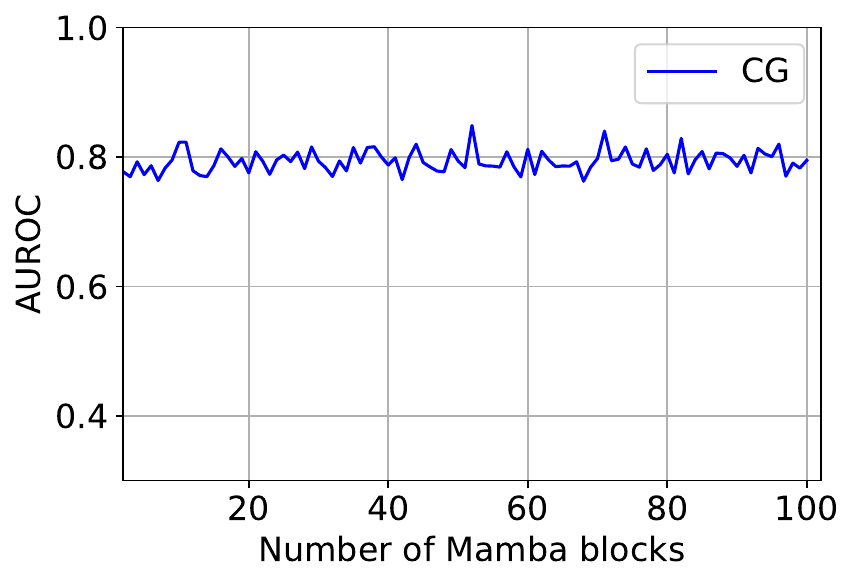} }}%
    \qquad
    \subfloat[Learnable parameters (CG)]{{\includegraphics[trim={0.25cm 0.25cm 0.3cm 0.25cm},clip,width=0.38\linewidth]{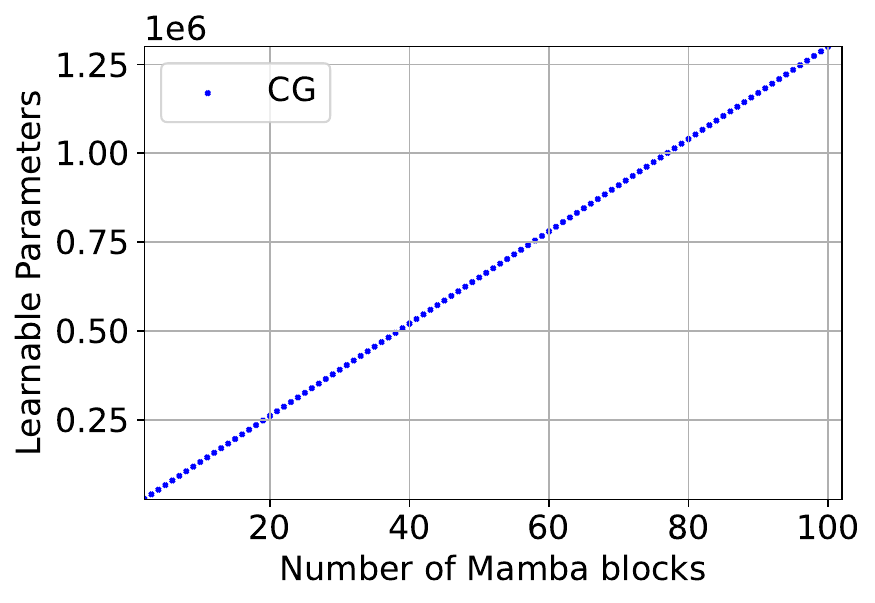} }}%
    \caption{Analysis on the of stacked residual Mamba blocks ($\mathcal{M}$).}%
    \label{fig:scaling_mamba}%
\end{figure}

Although we have achieved comparable or superior performance to current state-of-the-art methods with a default ${\mathcal{M}} = 1$ under regular supervised learning (see Table~\ref{tab:pub_dataset_results}), we also study the effect of scaling Mamba blocks via residual connections following ~\cite{he2016deep}. We stack Mamba blocks as in Figure~\ref{fig:method}, concatenating $\mathcal{M}=2$ up to 100 blocks, as shown in Equation~\ref{eq:res_connection}:
\begin{equation}
\label{eq:res_connection}
    h^{(i)}=Mamba_i(h^{(i-1)})+h^{(i-1)}.
\end{equation}
Here, $h^{(i)}$ is the hidden state from the $i$-th Mamba block, that is, $Mamba_i$, taking the prior block's hidden state $h^{(i-1)}$ as input. As seen in Figure~\ref{fig:scaling_mamba}, with increasing Mamba blocks, MambaTab retains comparable performance while the learnable parameters increase linearly on both CG and CB datasets. This demonstrates the Mamba block's information retention capacity. We observe that few Mamba blocks suffice for strong performance. Hence, we choose to use $\mathcal{M}=1$ by default.

\section{Future Scope}
Although we have evaluated our method on tabular datasets for classification, in the future we would like to incorporate our method for regression tasks as well on tabular data. Our method is flexible enough to incorporate regression tasks since we have kept the output layer open to predict real values. Therefore, our future research scope includes but is not limited to evaluating performance on different learning tasks.
\section{Conclusion}
This paper presents MambaTab, a plug-and-play method for learning tabular data. It uses Mamba, a state-space-model variant, as a building block to classify tabular data. MambaTab can effectively learn and predict in vanilla supervised learning, feature incremental learning settings, and adaptable to self-supervised learning. MambaTab demonstrates superior performance over current state-of-the-art deep learning and traditional machine learning-based baselines under supervised, feature incremental learning, and self-supervised learning on 8 public benchmark datasets. Remarkably, MambaTab occupies only a small fraction of memory in learnable parameter size compared to Transformer-based baselines for tabular data. Extensive results demonstrate MambaTab's efficacy, efficiency, and generalizability across diverse datasets for various tabular learning applications.
\section*{Acknowledgments}
We thank the creators of the public datasets and the authors of the baseline models and Mamba~\cite{gu2023mamba} for making these resources available for research.
This research is supported in part by the NSF under Grant IIS 2327113 and the NIH under Grants R21AG070909, P30AG072946, and R01HD101508-01.
\clearpage
\bibliographystyle{ieeetr}
\bibliography{bibliography}
\end{document}